\pdfoutput=1
\PassOptionsToPackage{dvipsnames}{xcolor}
\documentclass[11pt]{article}

\usepackage{acl}

\usepackage{times}
\usepackage{latexsym}

\usepackage[T1]{fontenc}

\usepackage[utf8]{inputenc}

\usepackage{microtype}

\usepackage{inconsolata}

\usepackage{paralist}
\usepackage{soul}
\usepackage{url}
\usepackage{graphicx}
\usepackage{amsmath}
\usepackage{amsthm}
\usepackage{booktabs}
\usepackage{algorithm}
\usepackage{algorithmic}
\usepackage[switch]{lineno}
\usepackage{multirow}
\usepackage{multicol}

\usepackage{listings}
\usepackage{amsmath,amsfonts,bm}
\usepackage{amssymb}

\def\eqref#1{equation~\ref{#1}}

\def\1{\bm{1}}

\DeclareMathAlphabet{\mathsfit}{\encodingdefault}{\sfdefault}{m}{sl}
\SetMathAlphabet{\mathsfit}{bold}{\encodingdefault}{\sfdefault}{bx}{n}

\usepackage[noabbrev,capitalize]{cleveref}

\crefname{equation}{equation}{equations}
\crefname{line}{line}{lines}
\crefname{section}{\S}{\S\S}
\Crefformat{section}{#2\S#1#3}
\crefrangeformat{section}{\S\S#3#1#4--#5#2#6}

\title{Competition-Level Problems are Effective LLM Evaluators}

\author{
Yiming Huang\textsuperscript{\rm 1}\thanks{\ \ Equal contribution.}\thanks{\ \ This work was done during their internship at MSRA.},
Zhenghao Lin\textsuperscript{\rm 2}\footnotemark[1]\footnotemark[2],
Xiao Liu\textsuperscript{\rm 1}\thanks{\ \ Corresponding authors.},
Yeyun Gong\textsuperscript{\rm 1}\footnotemark[3],
Shuai Lu\textsuperscript{\rm 1},
\\
\textbf{
Fangyu Lei,
Yaobo Liang\textsuperscript{\rm 1},
Yelong Shen\textsuperscript{\rm 3},
Chen Lin\textsuperscript{\rm 2}\footnotemark[3],
Nan Duan\textsuperscript{\rm 1},
Weizhu Chen\textsuperscript{\rm 3}\footnotemark[3]
}
\\
\textsuperscript{\rm 1}
Microsoft Research Asia,
\textsuperscript{\rm 2}
Xiamen University,
\textsuperscript{\rm 3}
Microsoft Azure AI
\\
\tt{\{xiaoliu2,yegong, wzchen\}@microsoft.com}, \tt{chenlin@xmu.edu.cn}
}

\begin{document}
\maketitle
\begin{abstract}
Large language models (LLMs) have demonstrated impressive reasoning capabilities, yet there is ongoing debate about these abilities and the potential data contamination problem recently.
This paper aims to evaluate the reasoning capacities of LLMs, specifically in solving recent competition-level programming problems in Codeforces, which are expert-crafted and unique, requiring deep understanding and robust reasoning skills.
We first provide a comprehensive evaluation of GPT-4's perceived zero-shot performance on this task, considering various aspects such as problems' release time, difficulties, and types of errors encountered.
Surprisingly, the perceived performance of GPT-4 has experienced a cliff like decline in problems after September 2021 consistently across all the difficulties and types of problems, which shows the potential data contamination, as well as the challenges for any existing LLM to solve unseen complex reasoning problems.
We further explore various approaches such as fine-tuning, Chain-of-Thought prompting and problem description simplification. Unfortunately, none of them is able to consistently mitigate the challenges.
Through our work, we emphasize the importance of this excellent data source for assessing the genuine reasoning capabilities of LLMs, and foster the development of LLMs with stronger reasoning abilities and better generalization in the future.
\end{abstract}


\section{Introduction}

The rise of LLMs has generated significant interest in the artificial intelligence community.
These models, notably GPT-4 \cite{openai2023gpt4}, have displayed impressive reasoning capabilities that are being harnessed in various fields \cite{bubeck2023sparks}.
However, questions\footnote{\url{https://twitter.com/keirp1/status/1724518513874739618}} have been raised about how to accurately evaluate the reasoning abilities of LLMs and the extent of data contamination issues \cite{mialon2023gaia,zhou2023don}.

\begin{figure}[t]
\centering
\includegraphics[width=0.48\textwidth]{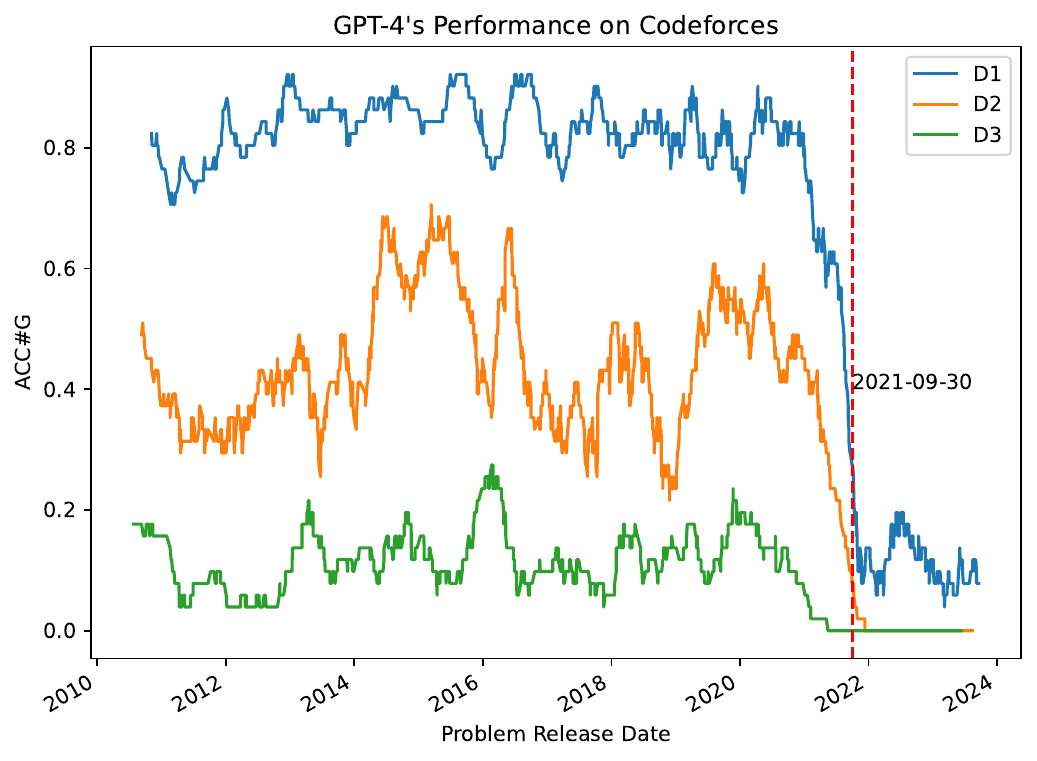}
\caption{The perceived zero-shot performance of GPT-4 sees a sharp decline on problems of varying difficulties (\textbf{D1}, \textbf{D2} and \textbf{D3} means easy, medium and difficult, respectively) in Codeforces after September 2021.}
\label{fig:ac_time}
\end{figure}

Regarding these issues, our study aims to assess the reasoning capabilities of LLMs through their ability to generate algorithms for solving competition-level programming problems.
These questions are meticulously crafted by experts to form rigorous competitions.
They possess high quality, are unique, and exhibit excellent discriminative ability.
The testing cases are also meticulously prepared.
This necessitates that LLMs deduce the solution from the presented scenario, which requires a thorough understanding of algorithms, combined reasoning and coding skills, and strong problem-solving abilities. These problems thus present a significant challenge to both human coders and LLMs. 
Consequently, competition-level programming problems serve as effective tools for evaluating the two issues previously discussed: they assess the reasoning abilities of LLMs and, due to the strict problem selection process in competitions, reduce the likelihood of data contamination in new problems.

Our research provides an in-depth analysis of the zero-shot performances of GPT-4 and other code LLMs on competition-level programming problems in Codeforces, considering factors such as release time, problem difficulty, and the types of errors encountered.
The main insights of our study include:
(1) GPT-4 performs significantly worse on programming problems released after September 2021, casting doubt on its actual reasoning abilities.
(2) GPT-4 shows limited capability to solve difficult problems, indicating potential weaknesses in complex problem-solving.
(3) GPT-4 struggles with the first test case, suggesting errors may stem from its understanding of the problem at hand.
(4) The related phenomenon can be also observed in other LLMs, indicating that insufficient reasoning ability may be a common problem.


To explore possible ways to enhance the zero-shot performances of these LLMs on competition-level programming problems, we investigate several methods to improve performance on unseen problems. These methods include supervised fine-tuning with code-specific LLMs, Chain-of-Thought prompting \cite{wei2022chain}, and problem statement simplification. Specifically, we fine-tuned CodeLlama\cite{codellama} and DeepSeek-Coder\cite{deepseek-coder}, which are specialized language models designed to handle programming-related tasks. However, none of these methods consistently mitigated the issue or resulted in noticeable performance improvements, particularly for more difficult problems. This finding indicates that difficult and unseen programming problems are effective evaluators of LLMs.

Overall, the primary contributions of this study lie in proposing and validating that recent competition-level programming problems serve as an excellent data source for assessing the genuine reasoning capabilities of LLMs. We aim to foster further research in this field by innovating new approaches to address the challenge of complex reasoning problems in LLMs and by establishing reliable evaluation benchmarks for LLMs that minimize the risk of data contamination.

\section{Problem Setup}
\begin{figure*}[ht]
\centering
\includegraphics[width=0.98\textwidth]{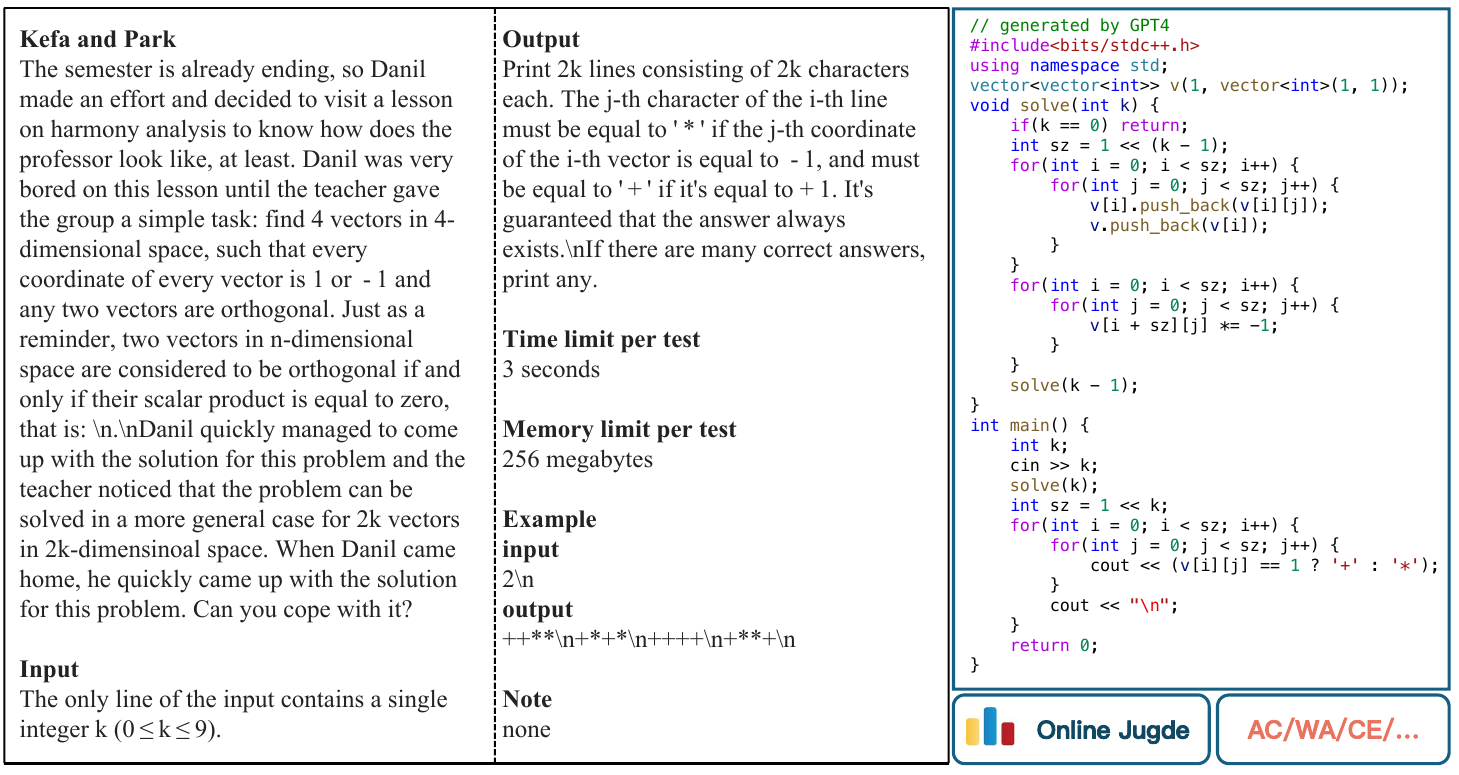}
\caption{The figure depicts the problem statement (left), comprising a problem set in a narrative context for participants to decipher, detailed input and output format specifications, and one or more example input-output pairs. In some cases, additional notes may be provided to assist competitors in understanding these example tests. This information is fed into the LLM, aiming to generate relevant code (right). The generated code is then submitted to an online judge for correctness evaluation.}
\label{fig:statement}
\end{figure*}

\label{sec:problem_definition}

\subsection{Competition-level Programming}
Competition-level programming presents a unique arena for testing and developing the reasoning abilities of AI models. In competitive programming, a problem typically consists of a narrative that sets the context, which models need to understand and convert into an algorithmic problem. The challenge lies in comprehending the narrative, identifying the underlying algorithmic issues, and implementing an efficient solution in programming languages such as C++ and Java.
Accepted programs must satisfy stringent testing conditions, including producing outputs that exactly match with test cases, executing within memory limits, and terminating within time constraints. In contrast to prior works \cite{humaneval,mbpp,multiple} focusing on basic coding abilities, competition-level programming problems require advanced reasoning and mathematical modeling skills, essential for AI.

Unlike the previous works that focused on LeetCode\footnote{\url{https://leetcode.com/}} \cite{bubeck2023sparks,shen2023pangu,sakib2023extending}, we follow AlphaCode \cite{alphacode} and choose Codeforces\footnote{\url{https://codeforces.com/}}.
Codeforces is universally acknowledged by competitors and enthusiasts in the International Collegiate Programming Competition\footnote{\url{https://icpc.global/}} (ICPC) and the International Olympiad in Informatics\footnote{\url{https://ioinformatics.org/}} (IOI) as a popular and suitable platform for developing abilities for algorithm contests. The regular contests hosted on this platform are crafted by human experts, and contain plenty of intricate programming problems and contests of high quality. These contests come with comprehensive and robust test cases and exhibit a low degree of problem overlap. The unique nature of these contest problems makes it highly unlikely to find similar content on the internet before the competition concludes. As a result, utilizing specific time-segmented datasets, like those from contests conducted post the introduction of LLMs, serves as an effective strategy to prevent data contamination \cite{zhou2023don}.

Codeforces employs the Elo rating system\footnote{\url{https://codeforces.com/blog/entry/102}} to rank its users and problems, categorizing all problems into 28 distinct difficulties, ranging from 800 to 3500.
Compared to commonly utilized metrics such as the ratio of accepted submissions or users, this difficulty rating mechanism is more suitable as it is based on the ranking and performance of the participants during the competition.\footnote{\url{https://codeforces.com/blog/entry/62865}} Hence, it is not subject to inaccuracies stemming from  temporal changes, repeated submissions, plagiarism, and other potential distortions. \looseness=-1




\subsection{Problem Definition}
\cref{fig:statement} presents an example of the problem statement $\pi$.
The input of LLM is instantiated with the problem statement $\pi$ and a prompt $\rho$ (like $\rho_1$ in \cref{table:prompts}).
The LLM $\Gamma$ takes the input to generate the code as $\alpha = \Gamma(\rho(\pi))$.
The generated code $\alpha$ is then evaluated by an online judge (OJ).
The evaluation process can be summarized in the following equation:

\begin{small}
\begin{align}
\mathrm{OJ}(\alpha) &= \mathrm{OJ}(\Gamma(\rho(\pi))) \in \text{\{AC, WA, CE, ...\}}  \nonumber
\end{align}  
\end{small}
In this equation, $\Gamma(\rho(\pi))$ denotes the code generated by LLM with the prompt $\rho$. The OJ platform then rigorously assesses the code for its correctness, computational efficiency, and adherence to specified input/output formats. With an extensive testing mechanism, the platform employs a wide range of test cases and hidden scenarios to ensure the code's robustness across diverse scenarios. The platform provides a spectrum of outcomes, $\mathrm{OJ}(\Gamma(\rho(\pi)))$, offering a holistic evaluation of the code's performance. This includes results such as Accepted (AC), Wrong Answer (WA), and Compilation Error (CE), among others. 

\subsection{Dataset Collection}
\label{sec:dataset_collection}


The dataset is compiled from the Codeforces website, extracting all publicly available problem statements from completed contests spanning February 2010 through November 2023.
For simplicity, problems requiring interaction, featuring non-standard input/output formats, or incompatible with C++ submission are excluded.
For detailed explanations, see \cref{sec:dataset_details}.

The analysis is confined to problems with difficulty levels ranging from 800 to 2400.
Based on their difficulty levels, the dataset is divided into three subsets: \textbf{D1} (800-1100 difficulty, 1683 problems), \textbf{D2} (1200-1600 difficulty, 1821 problems), and \textbf{D3} (1700-2400 difficulty, 1453 problems). These problems encompass more than 20 distinct categories of algorithms, as illustrated in \cref{table:tags}.
This diversity in problem types further enhances the comprehensiveness of the dataset and enables a comprehensive assessment of GPT-4's problem-solving abilities across a wide range of competition-level programming problems. 




\begin{table*}[t]
\small
\setlength\tabcolsep{7pt}
\centering
\begin{tabular}{c|ccc|ccc|ccc}
\hline
\multirow{2}{*}{\textbf{Metric}} & \multicolumn{3}{c|}{\textbf{D1}} & \multicolumn{3}{c|}{\textbf{D2}} & \multicolumn{3}{c}{\textbf{D3}} \\ \cline{2-10} 
 & \textbf{Time1} & \textbf{Time2} & \textbf{$\Delta$} & \textbf{Time1} & \textbf{Time2} & \textbf{$\Delta$} & \textbf{Time1} & \textbf{Time2} & \textbf{$\Delta$} \\ \hline
ACC\#G  & 81.42\% & 11.73\% & -69.69\% & 43.72\% & 0.00\% & -43.72\% & 11.41\% & 0.00\% & -11.41\% \\ 
pass@1 & 78.11\% & 10.54\% & -67.57\% & 42.38\% & 0.61\% & -41.77\% & 9.45\% & 0.18\% & -9.27\% \\    
ACC1\#1 & 78.05\% & 9.38\% & -68.68\% & 43.37\% & 0.00\% & -43.37\% & 8.48\% & 0.00\% & -8.48\% \\  
ACC1\#5 & 94.03\% & 20.09\% & -73.94\% & 69.02\% & 3.06\% & -65.96\% & 21.24\% & 0.88\% & -20.36\% \\     
ACC2\#5 & 88.34\% & 11.83\% & -76.51\% & 54.41\% & 0.00\% & -54.41\% & 12.36\% & 0.00\% & -12.36\% \\     
ACC3\#5 & 81.82\% & 9.38\% & -72.44\% & 42.42\% & 0.00\% & -42.42\% & 7.51\% & 0.00\% & -7.51\% \\    
\hline  
\end{tabular}
\caption{Performance of GPT-4 on different groups of problems: Time1 is the problems released from October 2010 to September 2021, and Time2 is the problems released from October 2021 to November 2023.}
\label{table:gpt4acc}
\end{table*}

\subsection{Experiment Details}
\begin{table}[t]
\small
\centering
\begin{tabular}{p{0.15\linewidth} | p{0.7\linewidth}} 
\hline
\textbf{Metric} & \textbf{Definition} \\ \hline    
ACC\#G & Proportion of accepted solutions using greedy sampling (temperature $t=0$). \\
ACC\#GN & The number of accepted solutions using greedy sampling (temperature $t=0$) within the sliding window. \\
ACC$k$\#$n$ & Proportion of problems with $k$ or more accepted solution with top-p samplings ($t=0.7$, $p=0.95$) for $n$ times. \\    
pass@$k$ & Estimated proportion of problems with at least one accepted solution. \\
\hline  
\end{tabular}
\caption{Definitions of evaluation metrics.} 
\label{table:evaluation}
\end{table}
In Codeforces, each problem belongs to a contest. Once the contest concludes, the problems are disclosed and become publicly submittable. Therefore, we submit the solutions to the contests that have concluded for evaluation.

To evaluate the results, we employ ACC\#G, ACC\#GN, ACC$k$\#$n$ and pass@$k$ as defined in \cref{table:evaluation}.
Specifically, for ACC$k$\#$n$ metric, we consider two settings: (1) $k=n=1$ and (2) $k \in \{1, 2, 3\}$ with $n=5$.
Following Codex \cite{chen2021evaluating}, pass@$k$ is computed as

\begin{small}
\begin{align}
\text{pass@$k$} &:= \mathop{\mathbb{E}}_{\text{Problems}} \left[ 1 - \frac{{\binom{n-c}{k}}} {\binom{n}{k}} \right]
\nonumber
\label{eq:estimator}
\end{align}
\end{small}
where $n$ is defined as the total number of generated samples per problem used for evaluation, and $c$ represents the count of correct samples out of $n$ that have successfully passed the unit tests.
Here we use $k=1$ and $n=5$ for pass@$k$.

In our experiment, we follow the zero-shot setting.
To select an appropriate prompt, we conduct preliminary experiments with three prompts, $\rho_1$, $\rho_2$, and $\rho_3$, as listed in \cref{table:prompts}, using two subsets of \textbf{D1} problems: one from February to December 2010 and the other from January to October 2023, each comprising approximately 100 problems.
The standard deviations are 0.015 and 0.018, respectively, indicating consistent performance.
Therefore, we choose $\rho_1$ as the prompt in the subsequent experiments.
Furthermore, we employ a sliding window approach for all temporal analyses to smooth the data, addressing the sporadic release schedule of the problems. This ensures a sufficient number of test problems at each time point, using a window size of 51 (25 before and 25 after the time point).

\section{Insights and Implications}
\label{sec:experiment}
\subsection{Faltering on Unseen Problems}
\label{sec:time}


In this section, we delve into a temporal analysis of GPT-4 (gpt-4-0613)'s performance on programming problems.
\cref{fig:ac_time} illustrates GPT-4's performance using the ACC\#G metric.
On problems released prior to September 2021, GPT-4 exhibits minor fluctuations at different levels across problems of varying difficulty.
However, for problems released after September 2021, a significant deviation from the normal fluctuation range is observed.
Interestingly, this timing coincides with the cut-off date for the GPT-4 training data as announced by OpenAI\footnote{\url{https://platform.openai.com/docs/models/gpt-4-and-gpt-4-turbo}}.
We then calculate the average performance on problems before and after September 2021, as shown in \cref{table:gpt4acc}.
On \textbf{D1} problems, GPT-4's ACC\#G plummets from 81.42\% to 11.73\%, marking a stark decrease of 69.69\%.
Even more strikingly, the ACC\#G drops to 0.00\% on both \textbf{D2} and \textbf{D3} problems, from 43.72\% and 11.41\%, respectively.
To validate the reliability of the conclusion, we also calculate the pass@1 metric, which exhibits a similar trend.
This observation raises thought-provoking questions about the severity of the drop and the correlation between the data cut-off date and the performance decline.

\begin{figure}[t]
\centering
\includegraphics[width=0.48\textwidth]{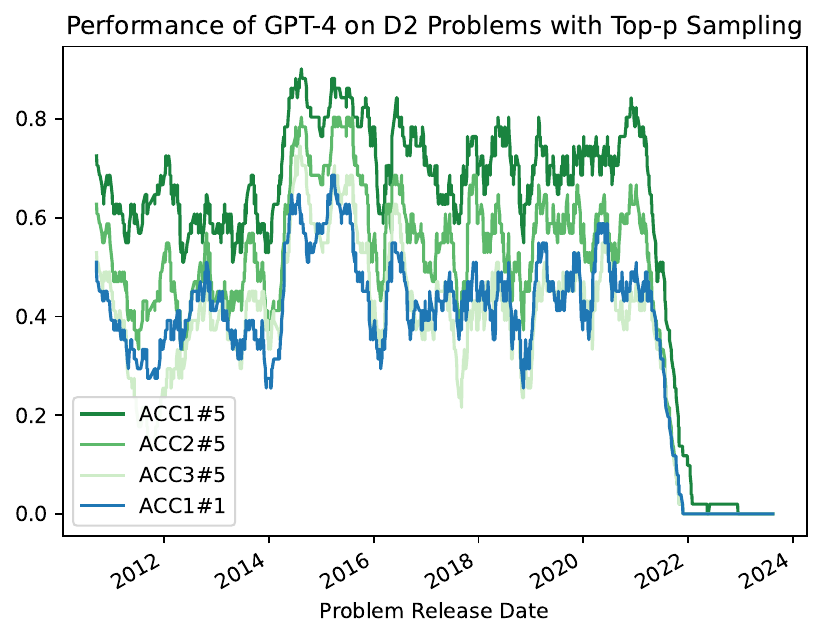}
\caption{Random sampling enhances the probability of generating correct solutions on previously encountered problems, but offers no assistance for unseen problems.}
\label{fig:sampling}
\end{figure}

To explore the model's potential to generate correct solutions, we perform random sampling multiple times and calculate the pass rate.
The average pass rate are shown in \cref{table:gpt4acc}.
As observed, multiple samplings can enhance the chances of generating a correct solution.
For instance, on the unseen simple \textbf{D1} problems, ACC1\#5 improved by 10.71\% compared to ACC1\#1.
However, across all problems, the performance gap before and after the cut-off date is more pronounced for ACC1\#5 than for both ACC1\#1 and ACC\#G.
\cref{fig:sampling} depicts the performance on \textbf{D2} problems over time.
A notable decline in performance metrics is observed around September 2021.
This observation underscores the challenges that LLMs, including the advanced GPT-4, face in addressing unseen programming problems without similar pretraining data.

The observed decline in performance on problems outside the model's training range may stem from limitations in reasoning and generalization. As highlighted by \citet{yadlowsky2023pretraining}, when confronted with problems beyond their pretraining data, transformer models exhibit various failure modes and their generalization abilities deteriorate, even for simple problems. Similarly, \citet{lu2023emergent} suggest that the exceptional abilities of large language models primarily stem from in-context learning, and do not necessarily reflect the emergence of reasoning abilities.  \looseness=-1

The observed performance drop on unseen problems raises serious questions about GPT-4's intrinsic reasoning and generalization capabilities.
This suggests a potential over-reliance on pattern recognition and reproduction from training, as opposed to grasping underlying principles and applying them to novel problems.
This observation aligns with recent debates on large models' data memorization tendencies \cite{carlini2023quantifying,yang2023rethinking}.
Therefore, future evaluations should prioritize the minimization of overlap between testing and training data to accurately assess a model's reasoning abilities, rather than simply its capacity for memorization. Furthermore, it's crucial to explore methods that enhance model generalization and reduce reliance on pre-training data. \looseness=-1

\begin{figure}[t]
\centering
\includegraphics[width=0.48\textwidth]{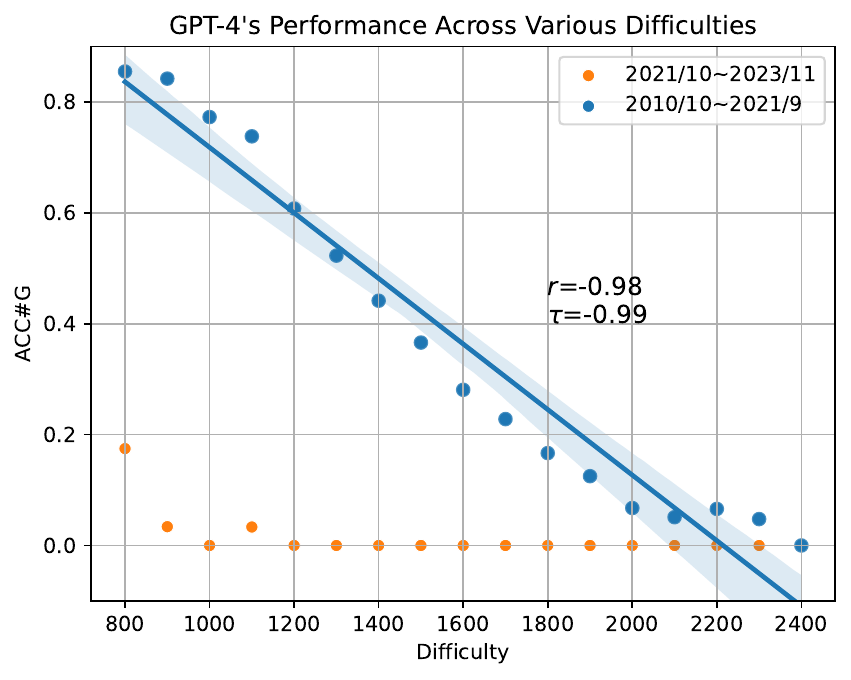}
\caption{For problems released before September 2021, GPT-4's ACC\#G showed a negative linear correlation with difficulty, followed by consistently poor performance afterwards.}
\label{fig:difficulity}
\end{figure}

\subsection{Limited Ability to Solve Difficult Problems}
\label{sec:Difficulity}
This section provides an analysis of performance in relation to the problem difficulty.
The results of ACC\#G for problems with different difficulties are reported for two distinct periods: from October 2010 to September 2021, and from October 2021 to November 2023, as illustrated in \cref{fig:difficulity}.

For the results from October 2010 to September 2021, we calculate Pearson correlation coefficient ($r=-0.97$) and the Kendall rank correlation coefficient ($\tau=-0.88$), which indicate strong linear correlations.
Notably, when the difficulty level reaches 2400 (indicating greater challenge than approximately 57\% of the problems on Codeforces), the ACC\#G drops to zero.
However, from October 2021 to November 2023, ACC\#G shows a dramatic decrease across all difficulty levels.

These findings reveal a significant limitation in the ability of GPT-4 to handle extremely complex problems. Despite its vast knowledge on code and algorithms, GPT-4 lacks of the competence in solving very challenging problems, particularly those with higher difficulty levels, even in the context of previously encountered problems. This indicates a potential area for further improvement and development in future iterations of the model.

\subsection{Struggling with The First Test Case}
\begin{figure}[t]
\centering
\includegraphics[width=0.48\textwidth]{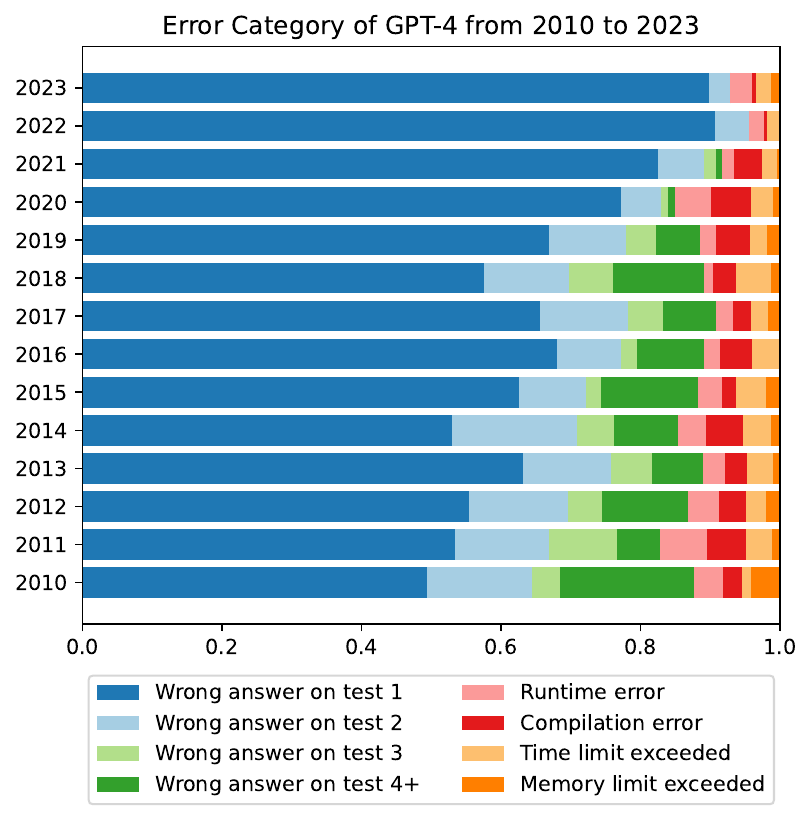}
\caption{Error categories in GPT-4's solutions on problems released from 2010 to 2023.}
\label{fig:error}
\end{figure}

In this section, we gather and analyze the errors returned by GPT-4 upon submission to the Codeforces website, as outlined in \cref{table:gpt4wrongtype}.
The most common error is "Wrong answer on test 1", which on average accounts for 70\% of the observed errors.
Test 1 is the first test case, which almost corresponds to or properly includes the example test case provided in the problem statement.
This suggests that the model often struggles at the very beginning of problem-solving, possibly due to difficulties in understanding the problem's requirements or generating a correct solution based on the given test case.
As depicted in \cref{fig:error}, there is a significant increase in the proportion of "Wrong answer on test 1" errors for problems released between 2021 and 2023.
This suggests that GPT-4 is more likely to face challenges in understanding and reasoning during at the onset of tackling unseen problems.

   
Other types of errors account for a smaller proportion, with an average of 10\%. They have shown little variation over time. This indicates that GPT-4 demonstrates strong fundamental code-writing capabilities of generating high-quality code.


\subsection{Similar Phenomenon of Other Code LLMs}
\begin{figure}[t]
    \centering
    \includegraphics[width=0.45\textwidth]{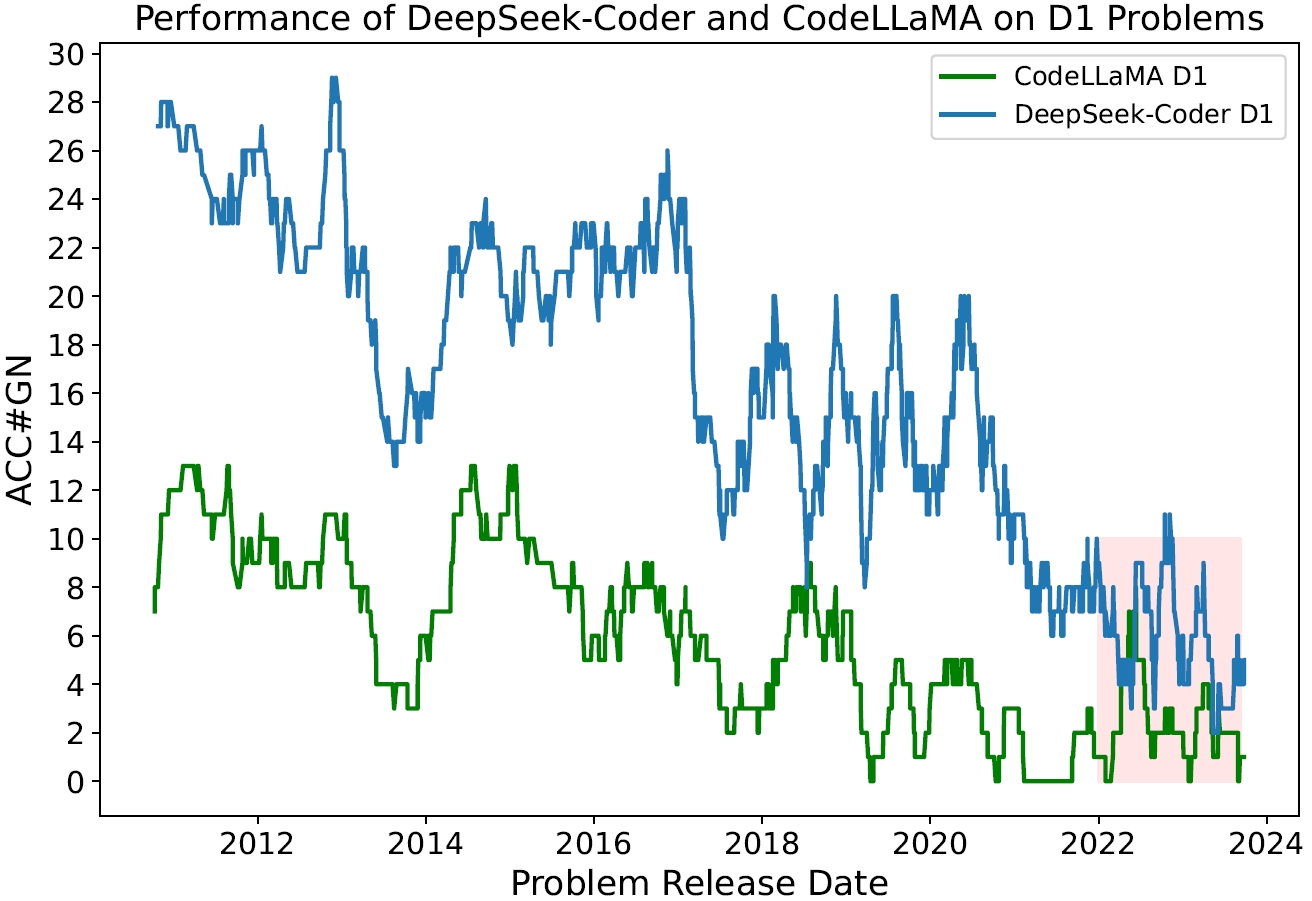}
    \caption{ACC\#GN of CodeLlama and DeepSeek-Coder on \textbf{D1} problems.}
    \label{fig:codellama_and_deepseek}
\end{figure}

\begin{table}[t]
\small
\center
\setlength\tabcolsep{2pt}
\begin{tabular}{c|c|c}
\hline
 \textbf{Problem Release Date} & \textbf{CodeLlama} & \textbf{DeepSeek-Coder} \\ \hline
Before 2023.3 & 10.30\% & 32.74\% \\
After 2023.3 & 4.52\% \textbf{(-5.78\%)} & 9.03\% \textbf{(-23.71\%)} \\ \hline
\end{tabular}
\caption{Comparison of ACC\#G between CodeLlama and DeepSeek-Coder on \textbf{D1} problems before and after March 2023.}
\label{tab:llamacode_and_deepseek_table}
\end{table}

We investigate whether the perceived performance degradation on unseen programming problems is observed for other popular code LLMs, such as CodeLlama-34B-Instruct~\cite{codellama} and DeepSeek-Coder-33B-Instruct~\cite{deepseek-coder}.

We conduct tests on CodeLlama and DeepSeek-Coder using \textbf{D1} problems, following the settings in \cref{sec:dataset_collection}, and the results are shown in \cref{fig:codellama_and_deepseek}.
The experimental results indicate that CodeLlama consistently underperforms compared to DeepSeek-Coder on \textbf{D1} problems.
Furthermore, the performance of DeepSeek-Coder on \textbf{D1} problems has been declining with the progression of the problem release date. The ACC\#GN of DeepSeek-Coder has declined to a level that is on par with CodeLlama when dealing with newly released problems, as highlighted in the red area of \cref{fig:codellama_and_deepseek}.





To precisely and intuitively detect this phenomenon, we calculate the ACC\#G of CodeLlama and DeepSeek-Coder on \textbf{D1} problems, both before and after March 2023, and present the results in \cref{tab:llamacode_and_deepseek_table}.
The results reveal a significant difference in the average accuracy of CodeLlama and DeepSeek-Coder before and after March 2023.
Regarding the magnitude of the decrease, DeepSeek-Coder, which previously exhibited superior performance, demonstrates a more pronounced decline, with acceptance rates falling below 10\% after March 2023.
Considering the release dates of CodeLlama and DeepSeek-Coder, we speculate that most of the programming problems after March 2023 are novel to them, which suggests that they also not be able to perform well on unseen programming problems like GPT4 does.
This finding indicates that a fundamental limitation of current code LLMs in generalizing effectively to complex reasoning tasks.
\subsection{Evaluation Hallucination of LLMs}
\begin{figure}[t]
    \centering
    \includegraphics[width=0.45\textwidth]{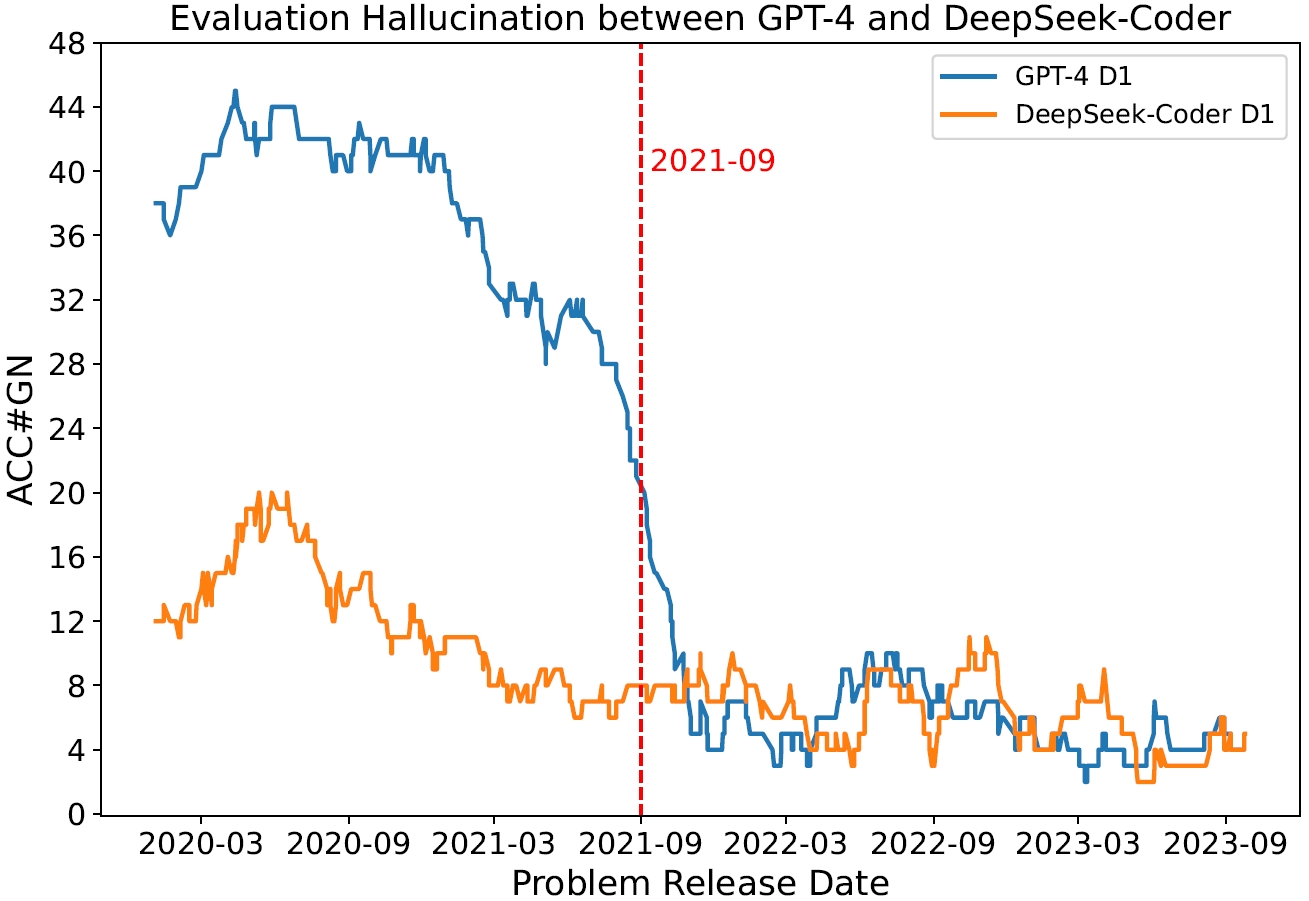}
    \caption{Comparison of ACC\#GN for GPT-4 and DeepSeek-Coder on \textbf{D1} problems after 2020.}
    \label{fig:gpt4_and_deepseek}
\end{figure}

\begin{table}[t]
\small
\center
\setlength\tabcolsep{2pt}
\begin{tabular}{c|c|c}
\hline
 \textbf{Model} & \textbf{2020.1-2021.9} & \textbf{2021.9-2023.10} \\ \hline
GPT-4 & 73.19\% \textbf{(+50.52\%)} & 11.53\% \textbf{(-0.97\%)} \\
DeepSeek-Coder & 22.67\% & 12.50\% \\ \hline
\end{tabular}
\caption{Comparison of ACC\#G between GPT-4 and DeepSeek-Coder over time intervals, on \textbf{D1} problems.}
\label{tab:gpt4_and_deepseek_table}
\end{table}


To further analyze the phenomenon, we compare GPT-4 with DeepSeek-Coder on \textbf{D1} problems as shown in \cref{fig:gpt4_and_deepseek} and \cref{tab:gpt4_and_deepseek_table}.

It is noteworthy that while GPT-4 surpasses DeepSeek-Coder in terms of performance on problems that were released prior to September 2021, an unexpected observation is that DeepSeek-Coder exhibits a performance that is on par with GPT-4 when it comes to tackling problems that were released after September 2021.
Considering the previous work \cite{yang2023rethinking, zhou2023don}, although GPT-4 may perform particularly well on some previously seen problems due to its powerful capacity, it cannot be well generalized on unseen programming problems, and its performance is not significantly different from DeepSeek-Coder, which is specifically trained for code. This phenomenon merits attention, which is termed as ``evaluation hallucination''.



Hence, a more equitable evaluation strategy would be to select evaluation sets that all the models have not previously encountered.
However, finding such data adhering to stringent conditions is challenging, as LLMs are typically pre-trained on extensive corpora containing diverse content, leading to the potential issue of data contamination.
Therefore, if we could devote more attention to the data source and timeline of the evaluation sets, such as the problems in Codeforces, it could potentially mitigate the effects of evaluation hallucination.

\section{One Step Forward}
\label{sec:future}

In this section, we explore some approaches to mitigate the poor performance on unseen problems.

\subsection{Fine-tuning}

\begin{figure}[t]
    \centering
    \includegraphics[width=0.45\textwidth]{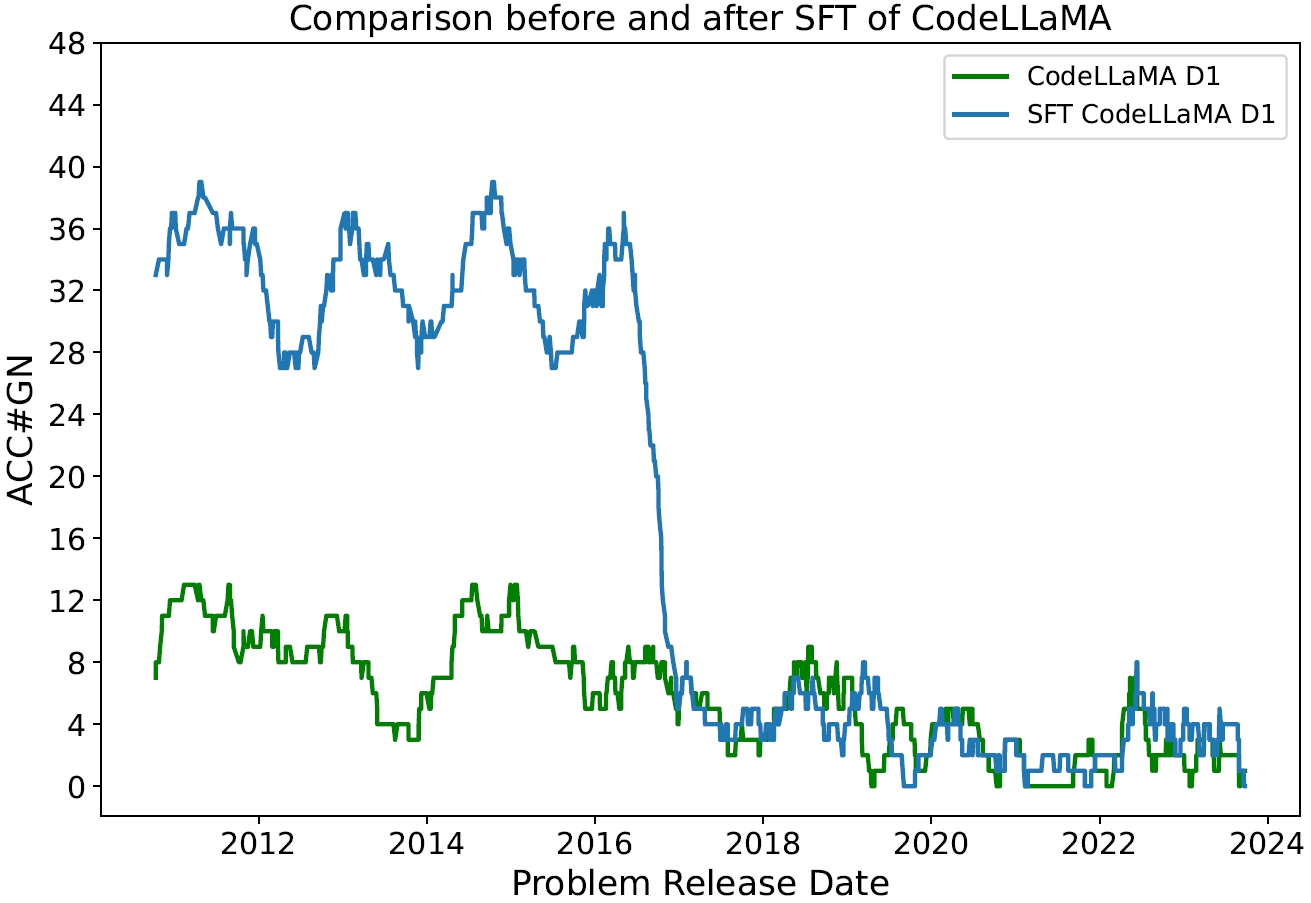}
    \caption{Comparison of ACC\#GN for CodeLlama on \textbf{D1} problems before and after fine-tuning.}
    \label{fig:sft_codellama}
\end{figure}

\begin{figure}[t]
    \centering
    \includegraphics[width=0.45\textwidth]{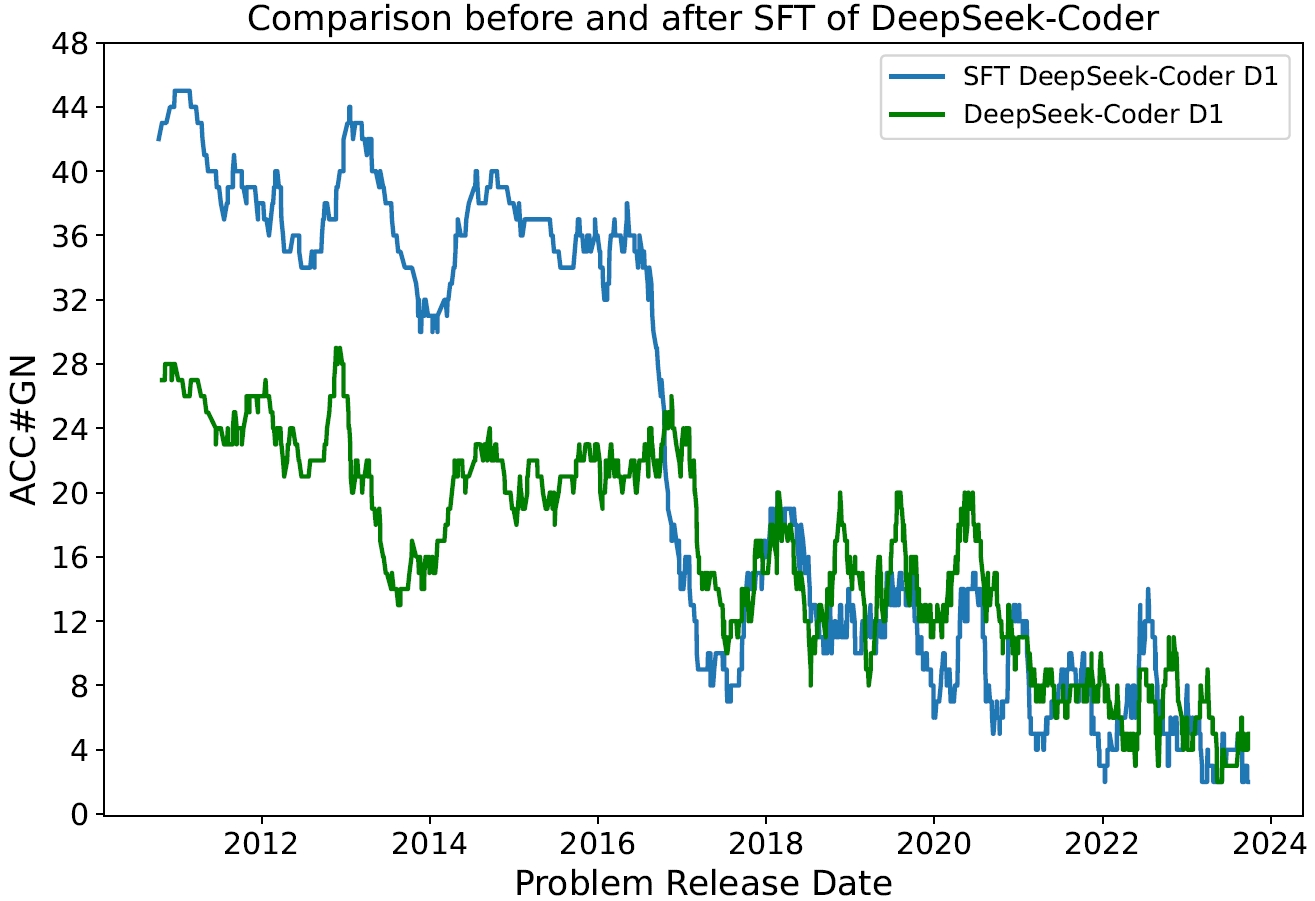}
    \caption{Comparison of ACC\#GN on \textbf{D1} problems before and after fine-tuning DeepSeek-Coder.}
    \label{fig:sft_deepseek}
\end{figure}

Fine-tuning is a commonly used method to improve performance on specific downstream tasks. In this study, we employ the Description2Code dataset \cite{Caballero_Description2Code_Dataset_2016} for fine-tuning. This dataset is compiled from three competitive programming websites: Codeforces, CodeChef, and HackerEarth, and contains problems published before 2017. CodeChef and HackerEarth, similar to Codeforces, host online coding competitions, and their problem sets are consistent in style and difficulty.

The dataset includes 2128 problems from Codeforces, 2435 problems from HackerEarth, and 3201 problems from CodeChef, totaling 7764 problems. However, due to some problems lacking corresponding C++ solutions, we retained a total of 7000 problems for our study. Each problem has approximately 10 C++ solutions, resulting in 70,000 pairs of input-output sequences. These sequences are used for fine-tuning both CodeLlama and DeepSeek-Coder in a supervised manner.

As shown in \cref{fig:sft_codellama} and \cref{fig:sft_deepseek}, we compare the performances of the models before and after fine-tuning on \textbf{D1} problems.
We observe that, even after fine-tuning with the same type of data, CodeLlama and DeepSeek-Coder do not exhibit improved performance on recent problems, particularly those post-2022.
The significant improvement in ACC\#GN before 2017 may result from the models recalling relevant or identical programming problems, rather than mastering the underlying reasoning logic, leading to their inability to adapt well to new programming challenges.
Therefore, simple fine-tuning does not effectively enhance the models' performance on new programming problems. \looseness=-1

\subsection{Chain-of-Thought Prompting}
\label{sec:cot}

In this section, we explore the application of Chain-of-Thought (CoT) prompting \cite{wei2022chain} to competition-level programming problems.
CoT involves prompting GPT-4 to generate an explanation of the algorithm before coding, denoted as $\rho_{cot}$ in \cref{table:prompts}.
We conduct experiments on both the \textbf{D1} and \textbf{D3} problems released after October 2021.
For \textbf{D1} problems, employing CoT increases the ACC\#G from 11.54\% to 16.21\%, demonstrating a noticeable improvement.
However, for \textbf{D3} problems, using CoT fails to yield any improvement, leaving the ACC\#G at 0.00\%.
This suggests that while CoT facilitates some improvement for simple \textbf{D1} problems, it is ineffective for the complex reasoning challenges presented by \textbf{D3} problems.

\subsection{In-Context Learning}

In this section, we enhance our experimental exploration into in-context learning by integrating both fixed demonstrations and retrieval-augmented demonstrations.

We use \textbf{D1} problems released before September 2021 as source dataset and those released after as test data. First, we apply the same method as delineated in \cref{sec:cot} to tackle the source data and validate them on Codeforces. We then retain accepted solutions, resulting in a collection of 1048 problem and CoT response pairs. A demonstration example is presented in \cref{tab:cot_demo}.

In the fixed demonstration experiment, we randomly select three problems to create 3-shot prompts. In the retrieval-augmented demonstration experiment, we first generate embeddings for the statements of the source and test data problems utilizing the OpenAI text-embedding-ada-002 model. We then identify the top three problems in the source data based on cosine similarity for each test problem, incorporating them as example demonstrations within the prompts.

The experimental results, summarized in \cref{table:rag}, show that the retrieval-augmented 3-shot method's accuracy is nearly identical to the 0-shot CoT, while the fixed 3-shot approach is even less effective. This may be due to the highly specialized nature of competitive programming problems, which makes finding valuable references challenging. Furthermore, the model may struggle to acquire problem-solving skills through context learning alone, and inappropriate demonstrations might lead to adverse effects.
\begin{table}[t]
\small
\center
\setlength\tabcolsep{2pt}
\begin{tabular}{c|c|c|c}
\hline
 \textbf{N-shot Prompt} &  \textbf{CoT} & \textbf{Retrieval} & \textbf{ACC\#G} \\
\hline
0-shot & No & No & 11.54\% \\
0-shot & Yes & No & 16.21\% \\
3-shot & Yes & No & 13.73\% \\
3-shot & Yes & Yes & 16.48\% \\
\hline
\end{tabular}
\caption{Accuracy of GPT-4 on \textbf{D1} problems released after September 2021 using different experimental setups}
\label{table:rag}
\end{table}

\subsection{Problem Statement Simplification}

Intuitively, even experienced programming competition competitors require time to understand problem statements.
Therefore, we conduct a simple experiment to assess whether comprehension of problem statements hinders LLMs' ability to excel at programming problems.
We first instruct GPT-4 to simplify the problem statement with $\rho_{sip}$ and then generate the code with $\rho_{sipgen}$ as shown in \cref{table:prompts}.
The results are also evaluated on both the \textbf{D1} and \textbf{D3} problems released after October 2021.
However, for \textbf{D1} problems, using the simplified problem statement even brings a slight decline in ACC\#G from 11.54\% to 11.14\%.
And the ACC\#G for \textbf{D3} problems still remains at 0.00\%.
Consequently, the challenge of genuinely improving the model's reasoning ability and enhancing its performance on unseen problems represents a significant direction for future research.



\section{Related Work}
\label{sec:related}

\paragraph{Code LLMs.}
Code intelligence is an important topic in AI research. Recently, code LLMs~\cite{zhang2023survey} have received widespread attention. Commercial LLMs~\cite{openai2023gpt4} have achieved tremendous success. Meanwhile, research on open-source code LLMs is also thriving, such as CodeLlama~\cite{codellama}, StarCoder~\cite{starcoder}, CodeGeeX~\cite{zheng2023codegeex}, CodeFuse~\cite{codefuse}, WizardCoder~\cite{wizardcode} and Lemur~\cite{lemur}.

\paragraph{Reasoning on Code.}
Programming competition is a specialized domain within the broader landscape of programming problems. Unlike simpler tasks on code, such as HumanEval~\cite{humaneval}, MBPP~\cite{mbpp}, MultiPL-E~\cite{multiple}, competition-level programming problems necessitate an advanced understanding of data structures, algorithms, and problem-solving techniques. 
Enabling models to solve human-designed algorithmic competition problems represents a meaningful research direction, as it reflects the models' integrated capabilities in reasoning, coding, and problem-solving.
AlphaCode~\cite{alphacode} simulate evaluations on 10 programming competitions on the Codeforces platform, which is the first work in this topic.
ALGO~\cite{zhang2023algo} can integrate with any existing code LLMs in a model-agnostic manner, enhancing its code generation performance. 

\paragraph{Reasoning on Other Subjects.}
Researchers have proposed many benchmarks requiring various reasoning skills, including commonsense reasoning~\citep{talmor2018commonsenseqa, geva2021did}, numerical reasoning~\citep{dua2019drop}, multi-hop reasoning~\citep{yang2018hotpotqa}, arithmetic reasoning~\citep{patel2021nlp,cobbe2021training}, structured reasoning~\citep{yu2018spider,lei2023s3eval}, inductive reasoning~\citep{sinha2019clutrr} and logical reasoning~\citep{yu2020reclor}.
LLMs are also widely used in scientific research in other fields~\citep{wang2023scientific}, such as physics~\citep{yeadon2023exploring}, chemistry~\citep{castro2023large, bran2023chemcrow}, etc.

\section{Conclusion}
\label{sec:conclusion}



In this study, we utilize competition-level programming problems from Codeforces to analyze the reasoning capabilities of LLMs.
We find a significant decrease in perceived performance of GPT-4 on unseen problems, consistent across a range of difficulties, problem types, and experimental settings.
This decrease highlights concerns of data contamination in benchmarks and the need for unseen tasks to properly assess LLMs' reasoning ability with complex challenges.
Our research also extends these insights to other open-source LLMs, revealing the common difficulties these models face with complex, previously unencountered reasoning tasks.
This is indicative of the LLMs' intrinsic limitations in reasoning.
As a primary probe, we explore several straightforward strategies, but none of them consistently mitigated the issues.
Through our work, we hope to emphasize the critical need for robust datasets to accurately evaluate LLMs' reasoning abilities and to inspire advancements in LLMs that demonstrate improved reasoning abilities. \looseness=-1

\section*{Limitations}
This study identifies expertly-designed, high-quality competition-level programming problems as effective evaluation data for evaluating LLMs. However, comparing to the existing benchmarks, the quantity of such problems is limited. Constructing uncontaminated, high-quality evaluation datasets and extending them to other tasks such as mathematics still poses a challenge to researchers. The identification and creation of such datasets are crucial for enhancing our understanding of the LLMs in complex reasoning tasks. We will endeavor to achieve this goal in our future work.


\bibliography{ref}

\newpage
\appendix
\section{More Results with Different Versions of GPT-4 APIs}
\begin{figure}[h]
\centering
\includegraphics[width=0.5\textwidth]{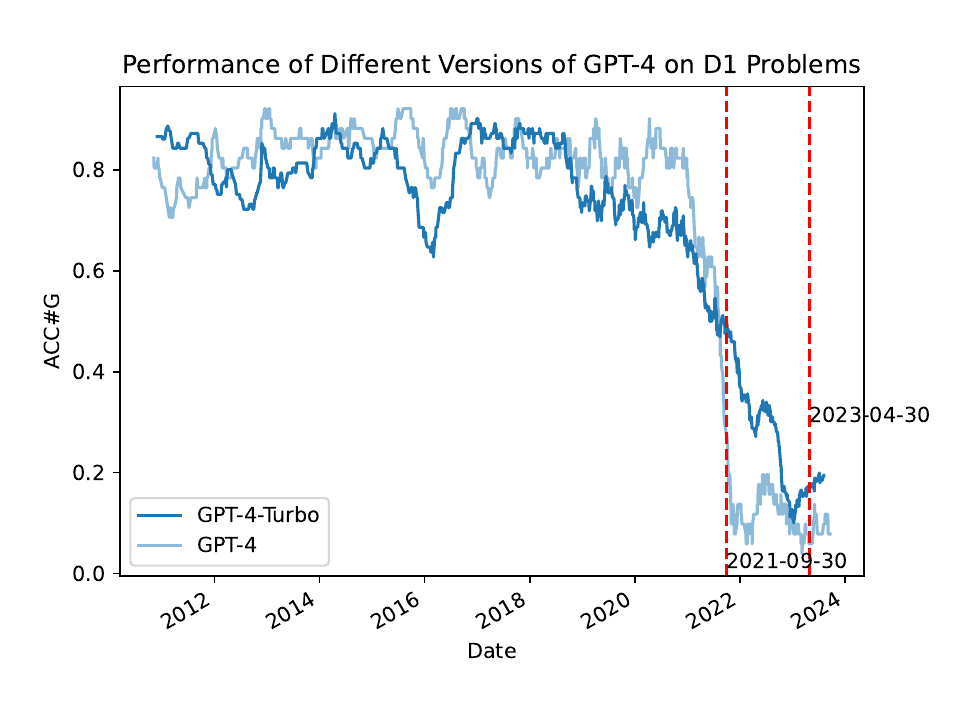}
\caption{Comparison of ACC\#G for GPT-4 and GPT-4-turbo on \textbf{D1} problems.}
\label{fig:gpt4-turbo}
\end{figure}

In this study, we conduct an evaluation of two distinct API versions: GPT-4 and GPT-4-turbo, to assess their performance on \textbf{D1} problems. The training data for GPT-4 extends up to September 2021, while that for GPT-4-turbo reaches up to April 2023. These evaluations are visually represented in \cref{fig:gpt4-turbo}. Upon analysis of the results, it is observed that on problems prior to September 2021, the GPT-4-turbo exhibits marginally inferior performance compared to GPT-4. Between September 2021 and April 2023, GPT-4-turbo outperforms GPT-4 on \textbf{D1} problems, reflecting the benefits of its more recent training data. Nonetheless, a decline in GPT-4's performance is observed for newer problems within this period, likely due to the scarcity of such recent data in its training set.

Nevertheless, when faced with problems emerging after April 2023—thus unencountered during their respective training periods—both APIs demonstrate a decline in performance, albeit GPT-4-turbo marginally outperforms GPT-4. Despite this relative improvement, the performance of GPT-4-turbo on problems post-April 2023 noticeably regresses when compared to its performance on problems covered by its training data. This finding is consistent with the conclusions drawn in the \cref{sec:time} "Faltering on Unseen Problems", which elucidates the challenges faced by these models when confronted with novel questions that extend beyond their training corpus.

\begin{table*}[h]
\small
\centering
\begin{tabular}{c|cccccccc}
\hline
\textbf{Year} & \textbf{WA1} & \textbf{WA2} & \textbf{WA3} & \textbf{WA4+} & \textbf{RE} & \textbf{CE} & \textbf{TLE} & \textbf{MLE} \\ \hline
2010 & 0.49 & 0.15 & 0.04 & 0.19 & 0.04 & 0.03 & 0.01 & 0.04 \\ 
2011 & 0.53 & 0.13 & 0.10 & 0.06 & 0.07 & 0.06 & 0.04 & 0.01 \\ 
2012 & 0.55 & 0.14 & 0.05 & 0.12 & 0.04 & 0.04 & 0.03 & 0.02 \\  
2013 & 0.63 & 0.13 & 0.06 & 0.07 & 0.03 & 0.03 & 0.04 & 0.01 \\ 
2014 & 0.53 & 0.18 & 0.05 & 0.09 & 0.04 & 0.05 & 0.04 & 0.01 \\  
2015 & 0.62 & 0.10 & 0.02 & 0.14 & 0.03 & 0.02 & 0.04 & 0.02 \\  
2016 & 0.68 & 0.09 & 0.02 & 0.10 & 0.02 & 0.05 & 0.04 & 0.00 \\  
2017 & 0.66 & 0.13 & 0.05 & 0.08 & 0.03 & 0.03 & 0.03 & 0.02 \\ 
2018 & 0.58 & 0.12 & 0.06 & 0.13 & 0.01 & 0.03 & 0.05 & 0.01 \\  
2019 & 0.67 & 0.11 & 0.04 & 0.06 & 0.02 & 0.05 & 0.02 & 0.02 \\  
2020 & 0.77 & 0.06 & 0.01 & 0.01 & 0.05 & 0.06 & 0.03 & 0.01 \\ 
2021 & 0.82 & 0.07 & 0.02 & 0.01 & 0.02 & 0.04 & 0.02 & 0.00 \\  
2022 & 0.91 & 0.05 & 0.00 & 0.00 & 0.02 & 0.00 & 0.02 & 0.00 \\  
2023 & 0.90 & 0.03 & 0.00 & 0.00 & 0.03 & 0.01 & 0.02 & 0.01 \\ 
Average & 0.70 & 0.10 & 0.03 & 0.06 & 0.03 & 0.03 & 0.03 & 0.01 \\ \hline
\end{tabular}
\caption{Error category of GPT-4 from 2010 to 2023. The abbreviations stand for: WA1, WA2, WA3, and WA4+ (Wrong Answers on Test 1, 2, 3, and 4 or above), RE (Runtime Error), CE (Compilation Error), TLE (Time Limit Exceeded), and MLE (Memory Limit Exceeded).}
\label{table:gpt4wrongtype}
\end{table*}

\section{Dataset Details}
\label{sec:dataset_details}
In the context of competitive programming challenges, a "non-standard input/output format" typically refers to a situation where the program's input and output are not provided through standard methods such as reading from standard input (stdin) or writing to standard output (stdout), which are the conventional ways for programs to receive and provide data during competitions. Instead, they might involve interacting with files, graphical user interfaces, or network connections, which are not commonly used in standard programming contests (like https://codeforces.com/problemset/problem/120/A). To filter out problems with non-standard input/output formats, we utilize metadata from the problem descriptions on Codeforces, which indicate whether a problem requires non-standard methods for input and output. By checking this information, we can automatically exclude such problems from our dataset to ensure the consistency of the test data.

Statistics of the types of problems in \textbf{D1}, \textbf{D2}, and \textbf{D3} are shown in \cref{table:tags}.
\begin{table*}[h]
\small
\centering
\begin{tabular}{c c | c c}
\hline
\textbf{Tag} & \textbf{\#Problems} & \textbf{Tag} & \textbf{\#Problems} \\ \hline
implementation & 1746 & greedy & 1441 \\ 
math & 1382 & brute force & 825 \\ 
constructive algorithms & 783 & dp & 577 \\ 
sortings & 514 & data structures & 391 \\ 
strings & 381 & binary search & 342 \\ 
number theory & 309 & graphs & 263 \\ 
dfs and similar & 244 & two pointers & 197 \\ 
combinatorics & 179 & bitmasks & 154 \\ 
geometry & 142 & trees & 137 \\ 
games & 87 & dsu & 84 \\ 
shortest paths & 66 & *special & 58 \\ 
probabilities & 52 & hashing & 48 \\ 
divide and conquer & 35 & flows & 24 \\ 
graph matchings & 22 & ternary search & 22 \\ 
matrices & 22 & expression parsing & 19 \\ 
string suffix structures & 10 & 2-sat & 7 \\ 
chinese remainder theorem & 5 & schedules & 4 \\ 
meet-in-the-middle & 4 & fft & 4 \\ \hline
\end{tabular}
\caption{Statistics of the types of problems in \textbf{D1}, \textbf{D2}, \textbf{D3}.}
\label{table:tags}
\end{table*}

\section{Prompt Details}
Prompts used in this study are shown in \cref{table:prompts}.

\newcommand{\cpp}{\textasciigrave \textasciigrave \textasciigrave cpp\textbackslash{}n...\textbackslash{}n\textasciigrave \textasciigrave \textasciigrave }
\newcommand{\n}{\textbackslash{}n}

\begin{table*}[t]
\centering
\small
\begin{tabular}{p{0.05\linewidth} p{0.8\linewidth}} 
\hline
$\rho_1$ & You are given a problem, you need to write a C++ solution and explain the algorithm.
\{$promblem\_name$\}

\{$promblem\_description$\}

Input specification: \{$input\_format$\}

Output specification: \{$output\_format$\}

Note: \{$note$\}

Memory limit: \{$memory\_limit$\}

Time limit: \{$time\_limit$\}

Example:

Input:

\{$input_i$\} 

Output:

\{$output_i$\}

Please provide a C++ code in \cpp  \\ \hline
$\rho_2$ & Read the problem, write a C++ solution and explain the algorithm.
\{$promblem\_name$\}: 
\{$promblem\_description$\}
Input specification is\{$input\_format$\}.  
Output specification is \{$output\_format$\}. 
Note that \{$note$\}. 
Memory limit is\{$memory\_limit$\}. 
Time limit is \{$time\_limit$\}.
Example {$i$} input is\{$input_i$\}. 
Example {$i$} output is\{$output_i$\}. 
Please provide a C++ code in \cpp  \\ \hline

$\rho_3$ & Finish the solution of this programming problem.

\{$promblem\_name$\}

\{$promblem\_description$\}

Input specification: \{$input\_format$\}

Output specification: \{$output\_format$\}

Note: \{$note$\}

Memory limit: \{$memory\_limit$\}

Time limit: \{$time\_limit$\}

Example:

Input:

\{$input_i$\} 

Output:

\{$output_i$\}

C++ code solution:

\textasciigrave \textasciigrave \textasciigrave cpp
\\ \hline

$\rho_{cot}$ & You are given an algorithm problem. First, provide a detailed explanation of the algorithm solution, including the logic behind it, the time and space complexity, and any important considerations or edge cases. Then, implement the solution in C++ code, ensuring it is clean, efficient, and well-commented.

\{$promblem\_name$\}

\{$promblem\_description$\}

Input specification: \{$input\_format$\}

Output specification: \{$output\_format$\}

Note: \{$note$\}

Memory limit: \{$memory\_limit$\}

Time limit: \{$time\_limit$\}

Example:

Input:

\{$input_i$\} 

Output:

\{$output_i$\}

Please provide a C++ code in \cpp  \\ \hline

$\rho_{sip}$ & Please extract the essential components from this algorithm problem for a C++ solution, removing any superfluous narrative or context.

\{$promblem\_name$\}

\{$promblem\_description$\}

Input specification: \{$input\_format$\}

Output specification: \{$output\_format$\}

Note: \{$note$\}
\\ \hline

$\rho_{sipgen}$ & You are given a problem, you need to write a C++ solution and explain the algorithm.

\{$promblem\_simlified$\}

Memory limit: \{$memory\_limit$\}

Time limit: \{$time\_limit$\}

Example:

Input:

\{$input_i$\} 

Output:

\{$output_i$\}

Please provide a C++ code in \cpp 
\\ \hline

\end{tabular}
\caption{Prompts used in this study.}
\label{table:prompts}
\end{table*}

\section{Case Study}
Some examples generated by GPT4 are shown in \cref{tab:example1,tab:example2,tab:example3,tab:cot_demo,tab:example4}.

\begin{table*}[t]
\centering
\begin{tabular}{p{0.2\linewidth} p{0.7\linewidth}} 
  \toprule
  \textbf{Problem Name}& \textbf{1890 B: Qingshan Loves Strings  ~~~~~~~~~~~~ \textcolor{green}{Accepted}}
  \\
  \midrule
  \textbf{Problem Statement}&
\textbf{Problem}
Qingshan has a string $s$, while Daniel has a string $t$. Both strings only contain $\texttt{0}$ and $\texttt{1}$.
A string $a$ of length $k$ is good if and only if $a_i \ne a_{i+1}$ for all $i=1,2,\ldots,k-1$. 
For example, $\texttt{1}$, $\texttt{101}$, $\texttt{0101}$ are good, while $\texttt{11}$, $\texttt{1001}$, $\texttt{001100}$ are not good.
Qingshan wants to make $s$ good. To do this, she can do the following operation any number of times (possibly, zero): insert $t$ to any position of $s$ (getting a new $s$). 
Please tell Qingshan if it is possible to make $s$ good.

\textbf{Input Format}
The input consists of multiple test cases. The first line contains a single integer $T$ ($1\le T\le 2000$) — the number of test cases. The description of the test cases follows.
The first line of each test case contains two integers $n$ and $m$ ($1 \le n,m \le 50$) — the length of the strings $s$ and $t$, respectively.
The second line of each test case contains a string $s$ of length $n$.
The third line of each test case contains a string $t$ of length $m$.
It is guaranteed that $s$ and $t$ only contain $\texttt{0}$ and $\texttt{1}$.

\textbf{Output Format}
For each test case, print "YES" (without quotes), if it is possible to make $s$ good, and "NO" (without quotes) otherwise.
You can print letters in any case (upper or lower).
 \\
  \midrule
  \textbf{Code}& 
  \begin{lstlisting}
#include<bits/stdc++.h>
using namespace std;

bool isGood(string s) {
    for(int i = 0; i < s.size() - 1; i++) {
        if(s[i] == s[i+1]) {
            return false;
        }
    }
    return true;
}

string solve(string s, string t) {
    if(isGood(s)) {
        return "YES";
    }
    if(!isGood(t)) {
        return "NO";
    }
    for(int i = 0; i < s.size() - 1; i++) {
        if(s[i] == s[i+1] && t[0] != s[i] && t[t.size()-1] != s[i+1]) {
            s.insert(i+1, t);
            i += t.size();
        }
    }
    if(isGood(s)) {
        return "YES";
    }
    return "NO";
}

int main() {
    int T;
    cin >> T;
    while(T--) {
        int n, m;
        cin >> n >> m;
        string s, t;
        cin >> s >> t;
        cout << solve(s, t) << endl;
    }
    return 0;
}
    
    \end{lstlisting}
\\
\bottomrule
\end{tabular}
\caption{An accepted example generated by GPT-4}
\label{tab:example1}
\end{table*}

\begin{table*}[t]
\centering
\begin{tabular}{p{0.2\linewidth} p{0.7\linewidth}} 
  \toprule
  \textbf{Problem Name}& \textbf{1893 A: Anonymous Informant  ~~~~~~~~~~~~ \textcolor{red}{Wrong answer on test 1}}
  \\
  \midrule
  \textbf{Problem Statement}&

\textbf{Problem}
You are given an array $b_1, b_2, \ldots, b_n$.
An anonymous informant has told you that the array $b$ was obtained as follows: initially, there existed an array $a_1, a_2, \ldots, a_n$, after which the following two-component operation was performed $k$ times:
A fixed point$^{\dagger}$ $x$ of the array $a$ was chosen. Then, the array $a$ was cyclically shifted to the left$^{\ddagger}$ exactly $x$ times.
As a result of $k$ such operations, the array $b_1, b_2, \ldots, b_n$ was obtained. You want to check if the words of the anonymous informant can be true or if they are guaranteed to be false.
$^{\dagger}$A number $x$ is called a fixed point of the array $a_1, a_2, \ldots, a_n$ if $1 \leq x \leq n$ and $a_x = x$.
$^{\ddagger}$A cyclic left shift of the array $a_1, a_2, \ldots, a_n$ is the array $a_2, \ldots, a_n, a_1$.

\textbf{Input Format}
Each test contains multiple test cases. The first line contains an integer $t$ ($1 \le t \le 10^4$) — the number of test cases. The description of the test cases follows.
The first line of each test case contains two integers $n, k$ ($1 \le n \le 2 \cdot 10^5$, $1 \le k \le 10^9$) — the length of the array $b$ and the number of operations performed.
The second line of each test case contains $n$ integers $b_1, b_2, \ldots, b_n$ ($1 \le b_i \le 10^9$) — the elements of the array $b$.
It is guaranteed that the sum of the values of $n$ for all test cases does not exceed $2 \cdot 10^5$.

\textbf{Output Format}
For each test case, output "Yes" if the words of the anonymous informant can be true, and "No" if they are guaranteed to be false.
 \\
  \midrule
  \textbf{Code}& 
  \begin{lstlisting}
#include<bits/stdc++.h>
using namespace std;

void solve() {
    int n, k;
    cin >> n >> k;
    vector<int> b(n);
    for(int i=0; i<n; i++) cin >> b[i];
    vector<int> prefix(n), suffix(n);
    prefix[0] = b[0];
    suffix[n-1] = b[n-1];
    for(int i=1; i<n; i++) prefix[i] = max(prefix[i-1], b[i]);
    for(int i=n-2; i>=0; i--) suffix[i] = max(suffix[i+1], b[i]);
    bool ok = true;
    for(int i=0; i<n; i++) {
        if(b[i] > max((i>0 ? prefix[i-1] : 0), (i<n-1 ? suffix[i+1] : 0))) {
            ok = false;
            break;
        }
        if(i > 0 && b[i] - b[i-1] > 1) {
            ok = false;
            break;
        }
    }
    if(b[0] != b[n-1]) ok = false;
    cout << (ok ? "Yes" : "No") << "\n";
}

int main() {
    ios_base::sync_with_stdio(false);
    cin.tie(NULL);
    int t;
    cin >> t;
    while(t--) solve();
    return 0;
}
    \end{lstlisting}
\\
\bottomrule
\end{tabular}
\caption{An wrong example generated by GPT-4}
\label{tab:example2}
\end{table*}

\begin{table*}[t]
\centering
\begin{tabular}{p{0.2\linewidth} p{0.7\linewidth}} 
  \toprule
  \textbf{Problem Name}& \textbf{1893 A: Anonymous Informant  ~~~~~~~~~~~~ \textcolor{red}{Compilation error}}
  \\
  \midrule
  \textbf{Problem Statement}&

\textbf{Problem}
You are given a set $S$, which contains the first $n$ positive integers: $1, 2, \ldots, n$.
You can perform the following operation on $S$ any number of times (possibly zero): 
Choose a positive integer $k$ where $1 \le k \le n$, such that there exists a multiple of $k$ in $S$. Then, delete the smallest multiple of $k$ from $S$. This operation requires a cost of $k$. 
You are given a set $T$, which is a subset of $S$. Find the minimum possible total cost of operations such that $S$ would be transformed into $T$. We can show that such a transformation is always possible.

\textbf{Input Format}
The first line of the input contains a single integer $t$ ($1 \le t \le 10\,000$) — the number of test cases. The description of the test cases follows.
The first line contains a single positive integer $n$ ($1 \le n \le 10^6$).
The second line of each test case contains a binary string of length $n$, describing the set $T$. The $i$-th character of the string is '1' if and only if $i$ is an element of $T$, and '0' otherwise.
It is guaranteed that the sum of $n$ over all test cases does not exceed $10^6$.

\textbf{Output Format}
For each test case, output one non-negative integer — the minimum possible total cost of operations such that $S$ would be transformed into $T$.
 \\
  \midrule
  \textbf{Code}& 
  \begin{lstlisting}
#include<bits/stdc++.h>
using namespace std;

int main() {
    ios_base::sync_with_stdio(false);
    cin.tie(NULL);

    int t;
    cin >> t;
    while(t--) {
        int n;
        cin >> n;
        string s;
        cin >> s;
        long long cost = 0, max_num = 0;
        for(int i = n - 1; i >= 0; i--) {
            if(s[i] == '1') {
                if(i + 1 > max_num) {
                    cost += i + 1;
                    max_num = i + 1;
                }
            } else {
                max_num = max(max_num, i + 1);
            }
        }
        cout << cost << "\n";
    }
    return 0;
}
    \end{lstlisting}
\\
\bottomrule
\end{tabular}
\caption{An example of compilation error generated by GPT-4}
\label{tab:example3}
\end{table*}

\begin{table*}[t]
\centering
\begin{tabular}{p{0.2\linewidth} p{0.7\linewidth}} 
  \toprule
  \textbf{Problem Name}& \textbf{451 A: Game With Sticks~~~~~~~~~~~~ \textcolor{green}{Accepted}}
  \\
  \midrule
  \textbf{Problem Statement}&
\textbf{Problem}
After winning gold and silver in IOI 2014, Akshat and Malvika want to have some fun. Now they are playing a game on a grid made of n horizontal and m vertical sticks. An intersection point is any point on the grid which is formed by the intersection of one horizontal stick and one vertical stick. In the grid shown below, n=3 and m=3. There are n+m=6 sticks in total (horizontal sticks are shown in red and vertical sticks are shown in green). There are n·m=9 intersection points, numbered from 1 to 9.

The rules of the game are very simple. The players move in turns. Akshat won gold, so he makes the first move. During his/her move, a player must choose any remaining intersection point and remove from the grid all sticks which pass through this point. A player will lose the game if he/she cannot make a move (i.e. there are no intersection points remaining on the grid at his/her move). Assume that both players play optimally. Who will win the game?

\textbf{Input Format}
The first line of input contains two space-separated integers, n and m ($1\le n$, $m\le 100$).

\textbf{Output Format}
Print a single line containing "Akshat" or "Malvika" (without the quotes), depending on the winner of the game.
 \\
  \midrule
  \textbf{Response}& 
The problem can be solved using a simple observation. The game is essentially about removing rows and columns from a grid. Each move removes one row and one column. Therefore, the game ends when there are no rows or columns left, i.e., when the minimum of n and m reaches zero. Since Akshat starts the game, if the minimum of n and m is even, Malvika will make the last move and win the game. Otherwise, Akshat will make the last move and win the game.

The time complexity of this solution is O(1) because it involves only a constant amount of computation. The space complexity is also O(1) because it uses only a constant amount of space.

The code first reads the values of n and m. It then checks if the minimum of n and m is even. If it is, it prints "Malvika". Otherwise, it prints "Akshat".

Here is the C++ code for the problem:
\begin{lstlisting}
#include<bits/stdc++.h>
using namespace std;

int main() {
    int n, m;
    cin >> n >> m;
    // If the minimum of n and m is even, Malvika wins. Otherwise, Akshat wins.
    if(min(n, m) % 2 == 0) {
        cout << "Malvika";
    } else {
        cout << "Akshat";
    }
    return 0;
}
\end{lstlisting}
\\
\bottomrule
\end{tabular}
\caption{An example generated by GPT-4 with CoT prompt.}
\label{tab:cot_demo}
\end{table*}

\begin{table*}[t]
\centering
\begin{tabular}{p{0.2\linewidth} p{0.7\linewidth}} 
  \toprule
  \textbf{Problem Name}& \textbf{1890 B: Qingshan Loves Strings}
  \\
  \midrule
  \textbf{Problem Statement}&
\textbf{Problem}
Qingshan has a string $s$, while Daniel has a string $t$. Both strings only contain $\texttt{0}$ and $\texttt{1}$.
A string $a$ of length $k$ is good if and only if $a_i \ne a_{i+1}$ for all $i=1,2,\ldots,k-1$. 
For example, $\texttt{1}$, $\texttt{101}$, $\texttt{0101}$ are good, while $\texttt{11}$, $\texttt{1001}$, $\texttt{001100}$ are not good.
Qingshan wants to make $s$ good. To do this, she can do the following operation any number of times (possibly, zero): insert $t$ to any position of $s$ (getting a new $s$). 
Please tell Qingshan if it is possible to make $s$ good.

\textbf{Input Format}
The input consists of multiple test cases. The first line contains a single integer $T$ ($1\le T\le 2000$) — the number of test cases. The description of the test cases follows.
The first line of each test case contains two integers $n$ and $m$ ($1 \le n,m \le 50$) — the length of the strings $s$ and $t$, respectively.
The second line of each test case contains a string $s$ of length $n$.
The third line of each test case contains a string $t$ of length $m$.
It is guaranteed that $s$ and $t$ only contain $\texttt{0}$ and $\texttt{1}$.

\textbf{Output Format}
For each test case, print "YES" (without quotes), if it is possible to make $s$ good, and "NO" (without quotes) otherwise.
You can print letters in any case (upper or lower).
 \\
  \midrule
  \textbf{Simplified \newline Statement}& 
You are given a problem, you need to write a C++ solution and explain the algorithm. 

- You are given two strings s and t, both containing only 0s and 1s.

- A string a of length k is considered good if no two consecutive characters are the same.

- You can insert string t at any position in string s any number of times.

- You need to determine if it is possible to make string s good.

- The input consists of multiple test cases. The first line contains an integer T ($1 \le T \le 2000$) — the number of test cases. Each test case starts with two integers n and m ($1 \le n, m \le 50$) — the lengths of the strings s and t, respectively. The next two lines contain the strings s and t.

- For each test case, print "YES" if it is possible to make s good, and "NO" otherwise. The output is case-insensitive.
\\
\bottomrule
\end{tabular}
\caption{An simplified example generated by GPT-4}
\label{tab:example4}
\end{table*}

\end{document}